\documentclass[fleqn,10pt]{olplainarticle}

\usepackage{url}
\usepackage{wrapfig}
\usepackage{booktabs}
\usepackage{multicol}
\usepackage{subcaption}
\graphicspath{{.}} \DeclareGraphicsExtensions{.pdf, .png}

\usepackage{bm}
\usepackage{amsmath}
\usepackage{amssymb}
\usepackage{mathtools}
\newcommand\SetSymbol[1][]{\nonscript\:#1\vert\allowbreak\nonscript\:\mathopen{}}
\providecommand\given{}
\DeclarePairedDelimiterX\Set[1]\{\}{\renewcommand\given{\SetSymbol[\delimsize]}#1}
\usepackage{siunitx}

\newcommand\ugmc[0]{\si[per-mode=symbol]{\micro\gram\per\meter\cubed}}

\newcommand\footer[1]{%
  \begingroup
  \renewcommand\thefootnote{}\footnote{#1}%
  \addtocounter{footnote}{-1}%
  \endgroup
}

\title{Lens functions for exploring UMAP Projections with Domain Knowledge}

\author[1]{Daniël Bot}
\author[2]{Jan Aerts}
\affil[1]{UHasselt, Data Science Institute (DSI) (e-mail: jelmer.bot@uhasselt.be)}
\affil[2]{KU Leuven, Augmented Intelligence for Data Analytics Lab, Department of Biosystems (e-mail:
jan.aerts@kuleuven.be)}

\begin{abstract}
    Dimensionality reduction algorithms are often used to visualise
    high-dimensional data. Previously, studies have used prior information to
    enhance or suppress expected patterns in projections. In this paper, we
    adapt such techniques for domain knowledge guided interactive exploration.
    Inspired by Mapper and STAD, we present three types of lens functions for
    UMAP, a state-of-the-art dimensionality reduction algorithm. Lens functions
    enable analysts to adapt projections to their questions, revealing otherwise
    hidden patterns. They filter the modelled connectivity to explore the
    interaction between manually selected features and the data's structure,
    creating configurable perspectives each potentially revealing new insights.
    The effectiveness of the lens functions is demonstrated in two use cases and
    their computational cost is analysed in a synthetic benchmark. Our
    implementation is available in an open-source Python package:
    \url{https://github.com/vda-lab/lensed_umap}.
\end{abstract}


\begin{document}
\flushbottom
\maketitle
\thispagestyle{empty}
\footer{This work is licensed under a Creative Commons
  Attribution-NonCommercial-NoDerivatives 4.0 International License. This work
  has been submitted to the IEEE for possible publication. Copyright may be
  transferred without notice, after which this version may no longer be
  accessible.}

\begin{multicols}{2}
\section{Introduction}%
\label{sec:introduction}%
Dimensionality reduction (DR) techniques are commonly used to visualise complex,
high-dimensional data in two or three dimensions~\citep{sacha2017interactivedr}.
While these visualisations provide a good overview, they may not contain all the
patterns an analyst expects to find. Patterns may not be visible due to errors
and distortions~\citep{nonato2019review} or can be hidden by a few influential
data attributes~\citep{fujiwara2023featurelearning}. Prior information and domain
knowledge have been incorporated in DR algorithms as constraints to make
embeddings better reflect analysts' expectations \citep{vu2022constraintdr}.
Generally, these techniques aim to emphasise or suppress known structures in the
embeddings.

In this paper, we use such constraints for domain-knowledge-guided exploration
instead. Our primary inspiration comes from topological data analysis algorithms
Mapper~\citep{singh2007mapper} and STAD~\citep{alcaide2015stad}. Both algorithms
support lens functions that highlight how particular features behave in
different parts of the data. Their key benefit is that they create configurable
perspectives, each potentially uncovering different insights. For example, lens
functions can be used to emphasise a feature of interest or to incorporate
additional signals.

The idea underpinning lens functions can be traced back to classical Morse
theory, which studies shape with a function that identifies points of interest
(e.g.,~\cite{biasotti2008reeb}). The Mapper algorithm is based closely on these
ideas, as it detects clusters in (overlapping) lens level sets and connects them
across level set boundaries to construct a network that approximates a Reeb
Graph~\citep{singh2007mapper}. The resulting structure summarises the relation
between the data's shape and the chosen lens function.

Lens functions can also be applied to modulate manifolds that describe
point-to-point connectivity. For example, STAD removes edges between points in
different lens level sets from a graph connecting all points closer than a
particular distance threshold~\citep{alcaide2015stad}. The resulting network does
not have the same formal properties as a Reeb graph but can be used to visualise
the same relations. In general, lens functions have three related effects: 1)
they separate similar observations with different lens values; thereby 2)
revealing distinct sub-populations with similar lens values; and 3) uncovering
how these sub-populations evolve over the lens function.

Visually, lens functions are particularly effective because they express
patterns by changing network connectivity and layout (i.e., position of the data
points on the screen). They modulate data point positions and change which data
points are pre-attentively perceived as a single group by the proximity Gestalt
law~\citep{ware2004gestalt}. Strictly speaking, many of the same patterns can
also be visualised by colouring data points. However, as the visual variable of
colour is less accurately perceived than
position~\citep{mackinlay1986visualvariables} they are not as easy to recognise
that way. This is also corroborated by the discovery of novel patterns in older
datasets using Mapper (e.g.,~\cite{lum2013mapper,nielson2015mapper}). 

The present paper proposes lens functions for UMAP, a state-of-the-art
dimensionality reduction algorithm~\citep{mcinnes2018umap}. We present three lens
types for UMAP models (see Fig.~\ref{fig:teaser}), bringing lens functionality
to UMAP in an accessible manner. Two case studies demonstrate the added value
lenses have for exploration workflows with UMAP: exploring data from multiple
perspectives, leading to different insights. In addition, we show the
computation scalability of these lens types by reporting compute times and
presenting a synthetic benchmark.

In summary, we make the following contributions:
\begin{itemize}
  \item Three lens types for UMAP models that adapt embeddings for answering
  questions using domain knowledge.
  \item Two use cases demonstrating exploration workflows using the lens types,
  and explaining in which scenarios each lens type is appropriate.
  \item A ready-to-use, open-source Python
  package\footnote{\url{https://github.com/vda-lab/lensed_umap}} implementing
  the proposed functionality and the demonstrated use cases.
\end{itemize}

\section{Related Work}%
\label{sec:related_work}
Our work has similarities to several other research topics. In this section, we
first describe three related research fields to contextualise our work: graph
signal processing, constraint dimensionality reduction, and visual analytics for
dimensionality reduction. Then, we introduce UMAP, the dimensionality reduction
algorithm on which we build.

\subsection{Graph Signal Processing}%
\label{sec:related_work:graph_signal_processing}
Lens values can be interpreted as a signal defined on a graph. The Graph Signal
Processing (GSP) field studies how such signals on graphs can be analysed
(e.g.,~\cite{shuman2013gsp,ortega2018gsp}). It provides tools that describe how
signals interact with the structure of a graph. In that sense, the field is
related to lens functions. However, where lens functions use a signal to change
the graph, GSP typically uses the graph to change or process the signal. For
example, GSP adapts the Fourier transform for signals on graphs, enabling
frequency-based transformations such as high-pass~\citep{sandryhaila2014highpass}
and low-pass filtering~\citep{zhu2012smooth},
translation~\citep{girault2015translation}, and
denoising~\citep{pang2015denoising}. The principles from GSP have also been used
to infer connectivity within graphs from the attributes present on the
nodes~\citep{dong2016graphlearning}. In addition, frequency and wavelet
coefficients can be used to detect interesting patterns within a
graph~\citep{mohan2014wavelets}, which could identify interesting lens
dimensions.

\subsection{Constraint Dimensionality Reduction}%
\label{sec:related_work:prior_knowledge}
Prior information and domain knowledge are typically incorporated in
dimensionality reduction algorithms as constraints~\citep{vu2022constraintdr}.
Several types of constraints are distinguished. Instance-level constraints apply
to individual points or relations between points. This type of constraint is
used to manipulate an embedding's layout~\citep{yang2006fixedpoints} or describe
which points should or should not be considered
similar~\citep{cevikalp2011pairwiseconstraints}. Dataset-level constraints apply
to datasets as a whole. For instance, to adapt feature
priorities~\citep{jeong2009ipca} or to incorporate class
labels~\citep{sugiyama2006fisher}, the data's hierarchy~\citep{hollt2019HSNE}, or
cluster shapes~\citep{machado2023sharp}. Recently, studies have also applied
constraints on the embedding's topological
structure~\citep{moor2020topologicalprior} to recover patterns that are known to
be in a
dataset~\citep{vandaele2022topologicalprior,heiter2023regularizedembeddings}.

Generally, these techniques aim to make the embedding reflect a known structure.
Prior information can also suppress known patterns to reveal other unexplained
patterns~\citep{kang2021conditionaltsne,heiter2023conditionaltsne}. Lens
functions have a different purpose. They use prior information to explore
datasets. They can, however, be explained in terms of constraints as a method
that introduces ``cannot-link'' relations between points that differ in lens
value. There is also a connection with \emph{multiview constraints} that
restrict embeddings to the variation shared between different views of the same
data items~\citep{vu2022constraintdr}, as lens functions may originate from such
different views.

\subsection{Visual Analytics for Dimensionality Reduction}%
\label{sec:related_work:enriched_projections}
Integrating dimensionality reduction algorithms in effective visual interfaces
for human analysts is an active research topic in the visual analytics field
(e.g.,~\cite{sacha2017interactivedr, nonato2019review}). We restrict our overview
to one broad task: interpreting patterns present in an embedding. Several
studies have designed visualisation systems for this purpose. For example,
t-viSNE explains patterns through data features that correlate most with
manually drawn polylines~\citep{chatzimparmpas2020ctvisne}. Colour has been used
to summarise which features are most stable or most extreme in value across a
projection~\citep{silva2015features, thijssen2023explainingprojections}. The
manifold's orientation around data points has been visualised by drawing
elliptical glyphs indicating each point's local linearised
variance~\citep{bian2020asubspaces}. Sequences, groups, and hierarchies have been
visualised directly in embeddings to let analysts summarise their data in these
terms and explain the structures in terms of the high-dimensional
data~\citep{eckelt2023relations}.

Lens functions have a similar goal to these techniques: they attempt to uncover
and explain patterns within a dataset that remain hidden when only a
dimensionality reduction's layout is shown. They differ from these techniques
because lens functions change the modelled structure---and thereby the produced
layout---rather than how the layout is shown. Lens functions can, therefore, be
combined with visualisation techniques that present additional information or
explain the patterns present within a layout.

A recent technique by Fujiwara et al~\citep{fujiwara2023featurelearning} is
perhaps the most related to our work. They generate multiple maximally distinct
projections from linear subspaces of a dataset to reveal patterns that are
unrecognisable when all features are projected. Our approach differs from theirs
by its human steerability and by the way the global structure is retained. Using
our method, analysts can target which features to inspect in the context of the
entire feature space rather than as a subspace.

\subsection{UMAP}%
\label{sec:related_work:UMAP}
\begin{figure*}[t!]
  \centering
  \begin{subfigure}{0.315\textwidth}
    \centering
    \includegraphics[width=\textwidth]{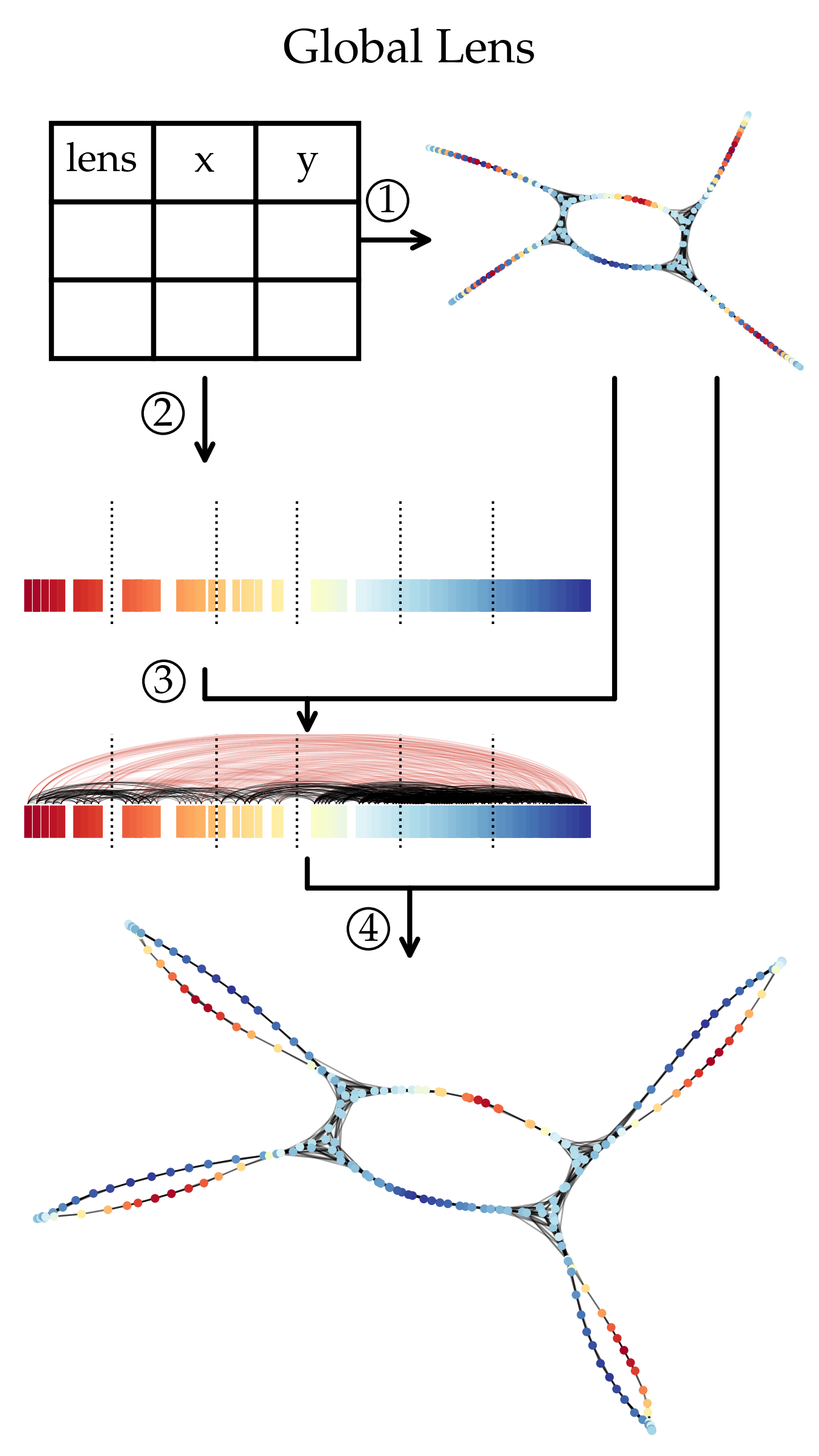}
    \caption{}%
    \label{fig:teaser:global-lens}
  \end{subfigure}\hfill
  \begin{subfigure}{0.315\textwidth}
    \centering
    \includegraphics[width=\textwidth]{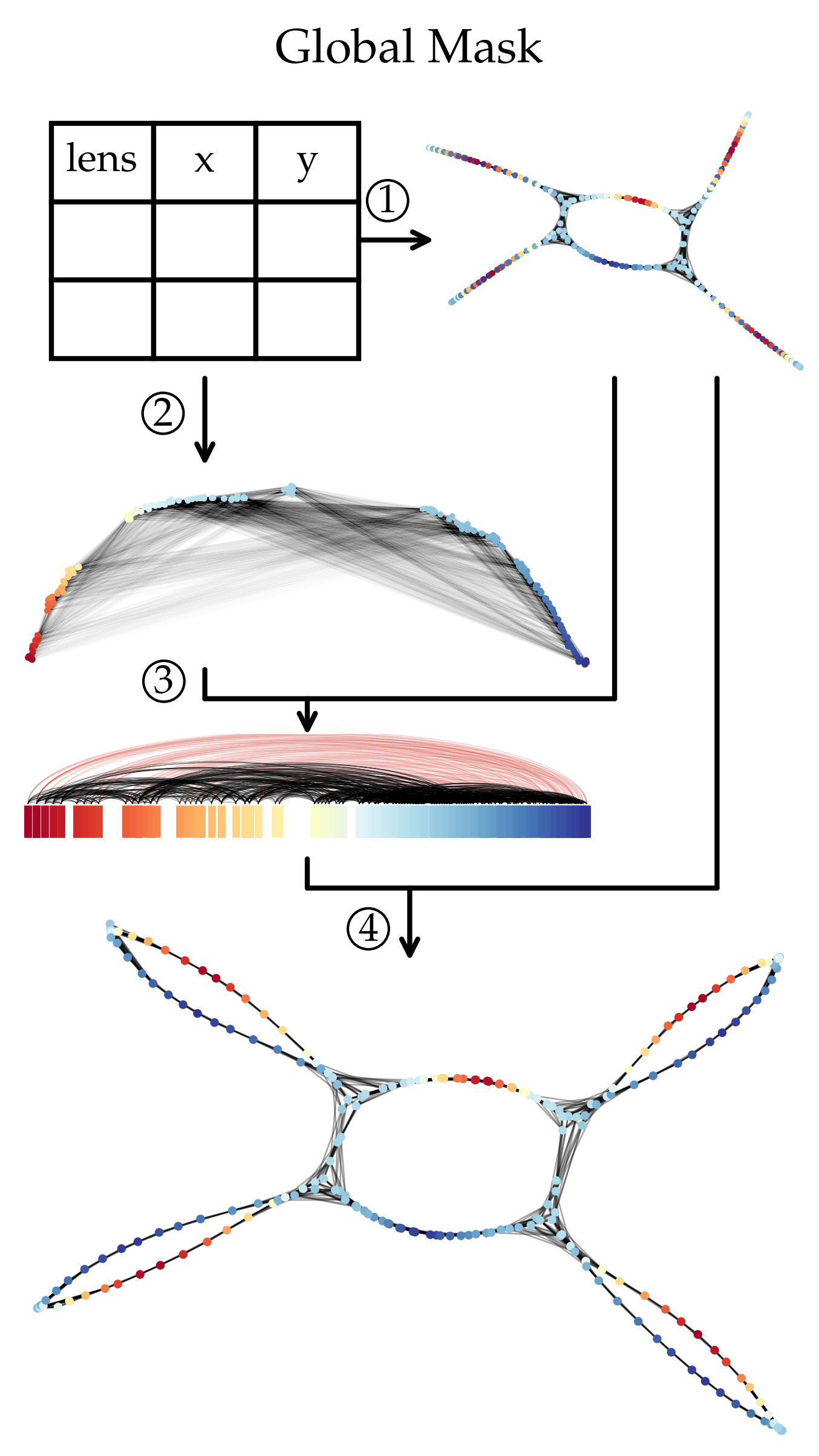}
    \caption{}%
    \label{fig:teaser:global-mask}
  \end{subfigure}\hfill
  \begin{subfigure}{0.315\textwidth}
    \centering
    \includegraphics[width=\textwidth]{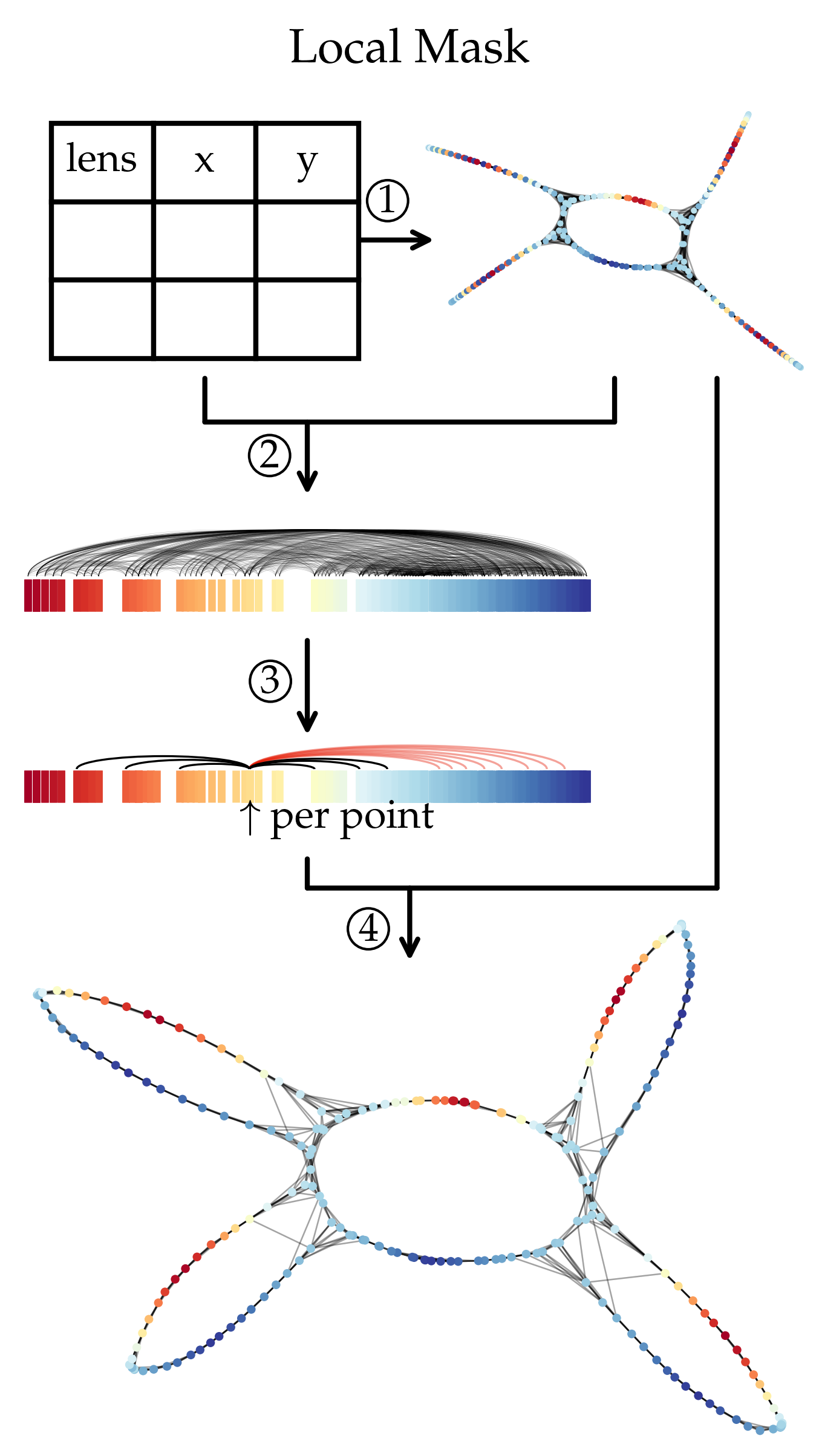}
    \caption{}%
    \label{fig:teaser:local-mask}
  \end{subfigure}
  \caption{Overview of the three lens types. All three lens types operate on an
     initial UMAP model, in this case constructed from a dataset with two
     spatial variables (1). The initial model does not reveal local lens extrema
     in its connectivity or layout, i.e., observations with low lens values
     (red) are connected and located near observations with high lens values
     (blue). The lens types filter the initial model's edges to separate
     observations that differ in the lens dimension. In the visualisations,
     edges that are kept are shown in black and edges that are removed are shown
     in red. How the lens types update the initial model differs: (\textbf{a})
     The global lens divides a single lens dimension---shown by horizontally
     ordered data points---into non-overlapping segments (2) and only keeps the
     initial model's edges between points in the same or neighbouring segments
     (3). (\textbf{b}) The global mask constructs a $k_{mask}$-nearest neighbour
     network over one or more lens dimensions (2) and only keeps the initial
     model's edges that also exist in the mask network (3). (\textbf{c}) The
     local mask computes the distance in one or more lens dimensions between
     points connected in the initial model (2) and only keeps the $k_{mask}$
     shortest ones for each point (3). All three lens types compute a layout for
     their updated model using the initial model's layout as starting point (4).
     The resulting embeddings reveal local extrema in the lens dimension.}%
  \label{fig:teaser}%
\end{figure*}
Uniform Manifold Approximation and Projection (UMAP) is a state-of-the-art
dimensionality reduction
algorithm~\citep{mcinnes2018umap,mcinnes2018implementation}. Together with
t-SNE~\citep{Vandermaaten2008tsne}, PBC~\citep{paulovich2006pbc}, and
IDMAP~\citep{minghim2006idmap}, UMAP produced the highest-rated embeddings in a
benchmark comparison considering multiple datasets and quality
metrics~\citep{espadoto2021survey}. Several studies have built upon UMAP since it
was first published, for example, to improve density
preservation~\citep{narayan2021densmap}, cluster
separation~\citep{dalmia2021connectivity}, and to embed multiple (overlapping)
datasets in an aligned manner~\citep{tariqul2022aligned}. In addition, a
specialised GPU implementation reaching interactive embedding speeds on large
datasets has been created~\citep{nolet2021rapids}. To our knowledge, the lens
functions we present here have not been implemented for UMAP before.

Because our work builds upon UMAP, explaining how the algorithm works is
relevant. Therefore, the remainder of this section presents a high-level
overview of the algorithm summarised from~\citep{mcinnes2018umap}. For a thorough
theoretical treatment of UMAP, we refer the reader to~\citep{mcinnes2018umap}.

\subsubsection{Approximating the Manifold}%
\label{sec:related_work:UMAP:manifold}
As its name suggests, UMAP works in two stages. The first approximates a
manifold along which the data is distributed uniformly. Let $X = \Set{\bm{x}_1,
..., \bm{x}_N}$ be the dataset and $d : X \times X \rightarrow
\mathbb{R}_{\ge0}$ a distance metric or dissimilarity function. Let $i_k$
indicate the index of $\bm{x}_i$'s $k$-th nearest neighbour. Then, the manifold
is computed from a directed $k$-nearest neighbour graph $G = (V, E, w)$, where
denotes $V$ the set of vertices, $E = \Set{(i, i_j) \given 1 \le j \le k, 1 \le
i \le N}$ is the set of edges, and $w(i, i_j)$ expresses the similarity between
$\bm{x}_{i_{j}}$ and $\bm{x}_i$ from $\bm{x}_i$'s perspective, accounting for
varying densities and ensuring each point is at least fully connected to its
closest neighbour. Finally, $G$ is symmetrised in a union operation that
combines the points' perspectives, interpreting $w(i, j)$ as the probability of
the edge existing in $E$.

\subsubsection{Projecting the Manifold}%
\label{sec:related_work:UMAP:projection}
UMAP's second stage typically functions as a graph layout algorithm for the
manifold graph. Formally, UMAP optimises an embedding into a user-defined space
to minimise the cross-entropy between the uncovered manifold $G$ and the
embedded points' manifold. Practically, UMAP employs a sampling-based stochastic
gradient descent strategy. The algorithm iterates for a pre-specified number of
epochs, sampling edges $(i, j)$ with a probability $w(i, j)$ to apply an
attraction force that increases the embedding similarity $\nu(i, j)$. The
high-dimensional similarity $w(i, j)$ is not used in the force computation once
an edge is selected. Similarly, a repulsion force decreases $\nu(i, k)$ for $m$
randomly selected vertices $k$. This negative sampling scheme assumes $w(i, k) =
0$, basing the applied force only on $\nu(i, k)$. These forces are applied using
a configurable learn-rate parameter that decays linearly to $0$ to improve
convergence.

This minimisation process is sensitive to the initialisation. Using an
initialisation that provides global structural information is essential to
preserve that information in the final
embedding~\citep{kobak2021initialisation}. The implementation's default spectral
initialisation performs that role but is negatively affected by disconnected
components and vertices~\citep{mcinnes2018umap}.

\section{Lensed UMAP}%
\label{sec:algorithm}
We present three types of lens functions for UMAP models that let analysts adapt
their projections to their questions. This section describes how they work in
detail. Table~\protect\ref{tab:lens-properties} summarises their overall
properties. All lens three types operate on a UMAP manifold $G = (V, E, w)$,
where $V$ is the set of vertices, $E$ is the set of edges, and $w(i, j)$ is the
edge weight acting as the probability data points $\bm{x}_i$ and $\bm{x}_j$ are
connected. The lens functions filter edges based on the lens dimensions and
(re-)project the manifold.

\subsection{Global Lens}%
\label{sec:algorithm:global_lens}
The global lens (Fig.~\ref{fig:teaser:global-lens}) is most similar to the
approaches used by Mapper~\citep{singh2007mapper} and
STAD~\citep{alcaide2015stad}. Like in STAD, the global lens first divides a
single lens dimension ($f: X \to \mathbb{R}$) into $k_{lens}$ non-overlapping
segments. Our implementation supports creating regularly spaced or balanced
segments encapsulating approximately the same number of points. Let $s_i$ be a
positive integer indicating point $\bm{x}_i$'s segment number (between $0$ and
$k_{lens}$). Then, edges are filtered, keeping only the edges between points
within the same or neighbouring segments:
\begin{equation}
  E_{global\;lens} = \Set{(i, j) \given (i, j) \in E, abs(s_i - s_j) \le 1 }.
\end{equation}
Extending this approach to circular lens domains is trivial by also allowing
edges between segments $0$ and $k_{lens}$. Multiple lens dimensions can be
combined by applying them in sequence.

This filtering approach differs from STAD in two ways. First, we keep empty lens
segments that reflect gaps in the lens dimension, which splits connected
components in the filtered graph. Second, we avoid needing a
community-detection-based post-processing step to maintain connectivity across
segment boundaries by allowing one boundary crossing.

The global lens can also be interpreted as a global lens distance threshold on
the edges in $E$. When regularly spaced segments are used, all edges with a lens
distance larger than one segment width are removed. With balanced segments, the
threshold varies with the lens distribution's density: for uncommon lens values,
the threshold is higher, and for common lens values, the threshold is lower.

The computational complexity of this lens type depends on the chosen
discretisation strategy. Computing regularly spaced segments has a complexity
linear with the number of points. The balanced segments require sorting the
points by their lens value. Filtering the edges has a complexity linear in the
number of edges.

\subsection{Global Mask}%
\label{sec:algorithm:global_mask}
The global mask (Fig.~\ref{fig:teaser:global-mask}) is similar to UMAP's
intersection functionality~\citep{mcinnes2018umap, mcinnes2018implementation}.
The global mask first computes a UMAP manifold $G_{mask} = (V, E_{mask},
w_{mask})$ over one or more lens dimensions ($f : X \to \mathbb{R}^m$) using
distance metric $d_{mask}: f(X) \times f(X) \to \mathbb{R}_{\ge0}$.
\begin{center}
  \captionof{table}{Summary of the lens types' properties. \textbf{Effect} indicates
    whether the lens type applies a global threshold or operates on the manifold
    locally. The other columns rank the lens types: \textbf{Tearing} indicates
    their tendency to split connected components, \textbf{Cost} indicates their
    computational cost, \textbf{Difficulty} indicates the intuitiveness of their
    parameters.}%
    \label{tab:lens-properties}
  \begin{tabular}{ c c c c c }
    \textbf{Lens type} & \textbf{Effect} & \textbf{Tearing} & \textbf{Cost} &
    \textbf{Difficulty} \\
    \midrule
    Global Lens        & Global          & Medium           & Low           &
    Medium              \\
    Global Mask        & Global          & High             & High          &
    High                \\
    Local Mask         & Local           & Low              & Medium        &
    Low                 \\
  \end{tabular}
  \vspace{1em}
\end{center}
Then, the initial model's edges are filtered, keeping only the edges that also
occur in $G_{mask}$:
\begin{equation}
  E_{global\;mask} = \Set{(i, j) \given (i, j) \in E, (i, j) \in E_{mask}}.
\end{equation}
The resulting graph is symmetrised, as the edges are undirected. Unlike UMAP's
intersection functionality, our filter does not adapt the edge weights $w(i,
j)$; it only removes edges that do not occur in the lens' manifold.

Like the global lens with balanced segments, the global mask can be interpreted
as a global lens distance threshold that varies with the lens distribution's
density. Here, that threshold is expressed as the number of neighbours in the
lens dimensions to keep. This number may need to be high when similar lens
values occur in multiple places along the manifold. Consequently, constructing
the lens manifold for large datasets can be quite expensive. The filter
operation is implemented like an element-wise sparse matrix multiplication,
which requires iterating over the edge union between $G$ and $G_{mask}$.

\subsection{Local Mask}%
\label{sec:algorithm:local_mask}
\begin{figure*}[!t]
  \centering
  \begin{subfigure}{0.298\textwidth}
    \includegraphics{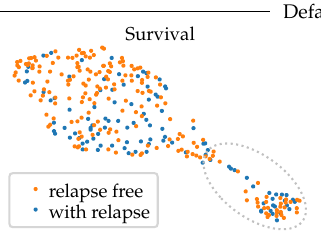}
    \caption{}%
    \label{fig:nki:base-survival}
  \end{subfigure}%
  \begin{subfigure}{0.351\textwidth}
    \includegraphics{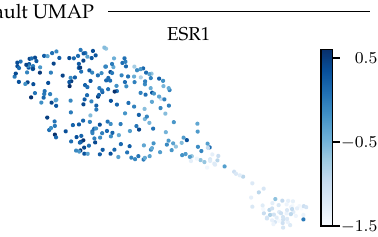}
    \caption{}%
    \label{fig:nki:base-esr1}
  \end{subfigure}%
  \begin{subfigure}{0.351\textwidth}
    \includegraphics{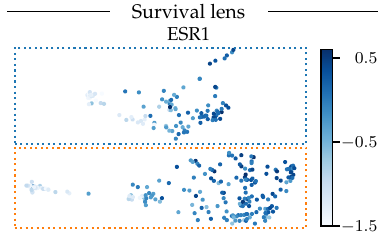}
    \caption{}%
    \label{fig:nki:survival-esr1}
  \end{subfigure}
  \par\medskip
  \begin{subfigure}{0.298\textwidth}
    \includegraphics{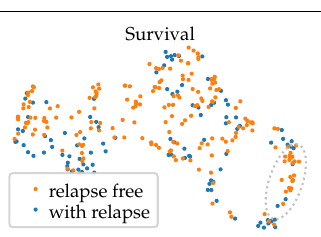}
    \caption{}%
    \label{fig:nki:csta-survival}
  \end{subfigure}%
  \begin{subfigure}{0.351\textwidth}
    \includegraphics{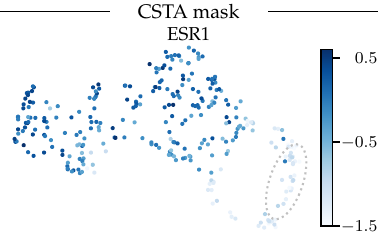}
    \caption{}%
    \label{fig:nki:csta-esr1}
  \end{subfigure}%
  \begin{subfigure}{0.351\textwidth}
    \includegraphics{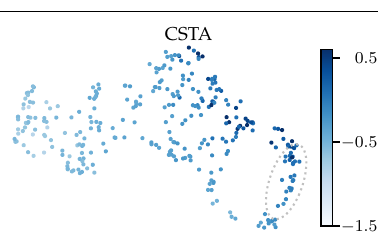}
    \caption{}%
    \label{fig:nki:csta-csta}
  \end{subfigure}%
  \caption{(Lensed) UMAP embeddings for the NKI
    dataset~\protect\citep{vantVeer2002nki}. (\textbf{a}) UMAP embedding
    (correlation distance, 30 nearest neighbours) coloured by survival.
    Contrasting patients within the grey dotted ellipse identifies the ESR1 gene
    (\textbf{b}). (\textbf{c}) A global lens with three segments separates
    patients by their survival state, indicated by the coloured rectangles.
    Contrasting patients by survival state within the low ESR1 community
    identifies the CSTA gene. A local mask (10 neighbours) over CSTA reveals how
    CSTA varies over the manifold (\textbf{d}) coloured by survival state,
    (\textbf{e}) coloured by ESR1, (\textbf{f}) coloured by CSTA. The grey
    dotted ellipse indicates a low ESR1, high CSTA region with an abundance of
    `relapse free' patients.}%
  \label{fig:nki}
\end{figure*}

The local mask (Fig.~\ref{fig:teaser:local-mask}) is most similar to a UMAP
manifold computed over the lens dimensions ($f: X \to \mathbb{R}^n$), where the
initial manifold $G$ prescribes the allowed edges. The local mask first computes
the lens dimension distance ($d_{mask} : f(X) \times f(X) \to
\mathbb{R}_{\ge0}$) for each edge in $G$. Then, let $r_i(\cdot)$ rank all edges
connected to $\bm{x}_i$ in $G$ by their (increasing) lens dimension distance,
such that $r_i(j)$ indicates edge $(i, j)$'s rank from $\bm{x}_i$'s perspective.
Then, the $k_{mask}$ shortest edges are kept for each point:
\begin{equation}
  E_{local\;mask} = \Set{(i, j) \given (i, j) \in E, r_i(j) < k_{mask}}.
\end{equation}
The resulting graph is symmetrised because the edges are undirected. As with the
previous lens types, the initial edge weights $w(i, j)$ are retained,
distinguishing this lens type from the previously mentioned UMAP manifold.

Unlike the global lens and global mask, this lens type cannot be reduced to a
global threshold. Instead, it operates in the context of each point, which
provides several benefits. Firstly, the number of neighbours parameter
$k_{mask}$ directly specifies how many edges should be kept for each point.
Secondly, the local lens is less likely to split connected components. Neither
gaps in the lens dimensions nor large lens value differences along the manifold
$G$ directly result in a tear, as each point is guaranteed to keep $k$ edges. On
the other hand, this also means that the local lens is less consistent in
removing large lens distance edges when few short lens distance edges connected
to a data point.

The computational complexity of this lens type depends on the initial manifold's
number of neighbours $k$ and the mask number of neighbours $k_{mask}$. Computing
each edge's lens distance has a linear complexity with the number of edges.
Then, finding each point's $k_{mask}$ closest lens-neighbours is, at worst, as
expensive as sorting each point's $k$ edge distances. Finally, constructing the
resulting graph has a complexity linear in the number of remaining edges.

\section{Use Cases}%
\label{sec:demo}
Lens functions help generating insights in exploratory data analyses by adapting
UMAP projections for particular questions. We present two use cases that
demonstrate the lens function in action and highlight how they provide benefits.
The first use case exhibits the similarities and differences with Mapper. The
second use case applies lensed UMAP to a larger dataset, and reports compute
times in a realistic setting. In addition, we present a synthetic benchmark to
investigates how the lens types' computational costs scale and are influenced by
their parameters. All timings were recorded on a computer with an AMD R7 7700
CPU and 32GB RAM.

\subsection{Breast Cancer Gene Expression}%
\label{sec:demo:nki}
This use case demonstrates the role of lens functions in an exploratory data
analysis using UMAP. We adapt a Mapper analysis~\citep{lum2013mapper} of the NKI
breast cancer dataset~\citep{vantVeer2002nki} that identified several interesting
genes in the Chemokine KEGG pathway. These genes distinguish patients with low
oestrogen receptor gene (ESR1) levels that relapse from those who remain
relapse-free. The strength of this exploration is that these genes can be
discovered without prior motivation to investigate low ESR1 patients. Instead,
visualising the networks raises the question of why particular sub-groups
differ, leading to the insights.

\subsubsection{Data and Pre-Processing}%
The data (obtained from~\cite{nkidata}) was pre-processed
following~\cite{lum2013mapper}. Specifically, we removed the rows and columns
with the 5\% most missing values. The remaining missing values were imputed
using their observation's $5$-nearest neighbours. Finally, we extracted the
$1553$ genes with the highest variance.

\subsubsection{Exploration Steps}%
The exploration starts by constructing a UMAP model using the correlation
distance and $30$ nearest neighbours, shown in
Fig.~\ref{fig:nki:base-survival}. The resulting embedding contains one larger
and one smaller community connected by a few data points, raising the question
of which genes differ between these communities. A Kolmogorov-Smirnov test
comparing expression values between the selected small community (grey-dotted
ellipse) and the other non-selected data points identified the ESR1 gene as
significantly different (p$<$0.01, D=0.90) (Fig.~\ref{fig:nki:base-esr1}).
This gene is relevant to a domain expert because low ESR1 expression has been
linked to poor prognoses (as cited in~\cite{lum2013mapper}).

At this point, a domain expert might wonder why there does not appear to be an
abundance of patients who relapse in this low ESR1 community. One way to explore
this question is to visually separate the patients by their survival state. A
global lens with three regular segments removes all connections between the two
groups. Fig.~\ref{fig:nki:survival-esr1} shows the resulting embedding, where
the two disconnected components were positioned below each other in a
post-processing step. The orange and blue rectangles indicate the `relapse-free'
and `with relapse' patients, respectively. This new embedding makes it easier to
see colour differences between the two groups.

The low ESR1 community can also be explored by comparing gene expressions
between its `relapse-free' and `with relapse' patients. The CSTA gene (among
others) was significantly different in a Kolmogorov-Smirnov test comparing these
groups (p$<$0.01, D=0.51). A local mask over CSTA reducing the model's
connectivity to $10$ nearest neighbours was applied to the original UMAP model
to investigate how CSTA behaves across the model.
Fig.~\ref{fig:nki:csta-survival}-\ref{fig:nki:csta-csta} show the resulting
embedding coloured by survival state, ESR1, and CSTA. In these figures, the low
ESR1 community is transformed into a loop along which CSTA increases from the
lower left to the upper right. A grey-dotted ellipse indicates a region with
many `relapse-free' patients. Patients in this region have low ESR1 but high
CSTA expressions, indicating CSTA expression correlates with survival state for
patients with low ESR1 expression.

The Chemokine genes identified by~\cite{lum2013mapper} also differed
significantly in the Kolmogorov-Smirnov test comparing survival state within the
low ESR1 community (p$<$0.05, D=0.36). This finding indicates that these genes
could be discovered using lensed UMAP. We chose to illustrate the local mask
with CSTA because its effect was stronger with our selections and
pre-processing. Also, note that we did not need an eccentricity lens to explore
the data in this use case. It is, however, possible to re-create the Y-shape
found by~\cite{lum2013mapper} with lensed UMAP using such a lens.

\subsection{Air-Quality}%
\label{sec:demo:air}
\begin{figure*}[t!]
  \centering
  \begin{subfigure}{0.315\textwidth}
    \centering
    \includegraphics{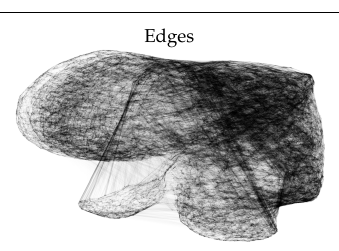}
    \caption{}%
    \label{fig:air:base-edges}
  \end{subfigure}%
  \begin{subfigure}{0.370\textwidth}
    \centering
    \includegraphics{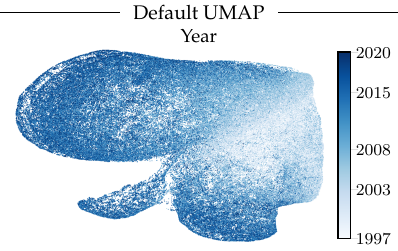}
    \caption{}%
    \label{fig:air:base-year}
  \end{subfigure}%
  \begin{subfigure}{0.315\textwidth}
    \centering
    \includegraphics{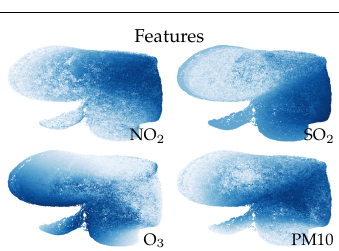}
    \caption{}%
    \label{fig:air:base-features}
  \end{subfigure}
  \par\medskip
  \begin{subfigure}{0.315\textwidth}
    \centering
    \includegraphics{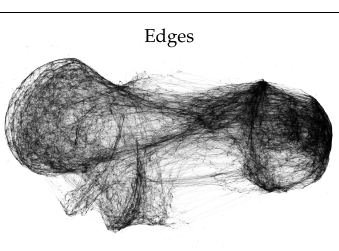}
    \caption{}%
    \label{fig:air:year-edges}
  \end{subfigure}%
  \begin{subfigure}{0.370\textwidth}
    \centering
    \includegraphics{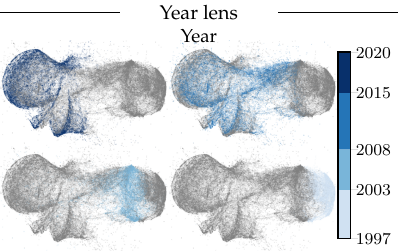}
    \caption{}%
    \label{fig:air:year-year}
  \end{subfigure}%
  \begin{subfigure}{0.315\textwidth}
    \centering
    \includegraphics{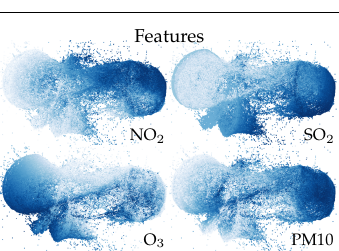}
    \caption{}%
    \label{fig:air:year-features}
  \end{subfigure}
  \par\medskip
  \begin{subfigure}{0.315\textwidth}
    \centering
    \includegraphics{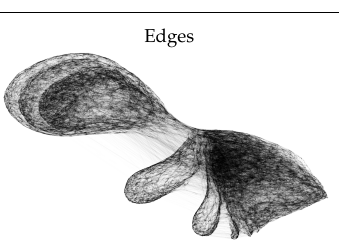}
    \caption{}%
    \label{fig:air:sulfer-edges}
  \end{subfigure}%
  \begin{subfigure}{0.370\textwidth}
    \centering
    \includegraphics{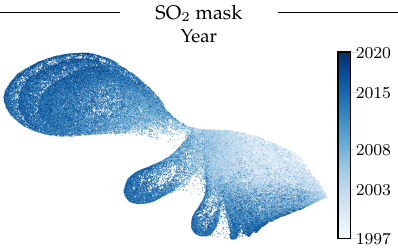}
    \caption{}%
    \label{fig:air:sulfer-sulfer}
  \end{subfigure}%
  \begin{subfigure}{0.315\textwidth}
    \centering
    \includegraphics{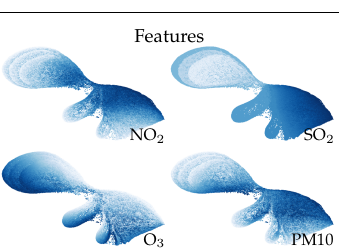}
    \caption{}%
    \label{fig:air:sulfer-features}
  \end{subfigure}
  \caption{(Lensed) UMAP embeddings for the Air Quality
    dataset~\protect\citep{airdata}. (\textbf{a})-(\textbf{c}) Default UMAP
    embedding (cosine distance, 50-nearest neighbours) shown by the model's
    edges and points coloured by year and features, respectively.
    (\textbf{d})-(\textbf{f}) The embedding after applying a global lens over
    the year dimensions (24 regular segments), drawn as before.
    (\textbf{g}-\textbf{i}) The embedding after applying a local mask (20
    neighbours) over the SO$_2$ dimension, drawn as before.}%
    \label{fig:air}
\end{figure*}
\begin{figure*}[t!]
  \centering
  \begin{subfigure}{0.333\textwidth}
    \centering
    \includegraphics{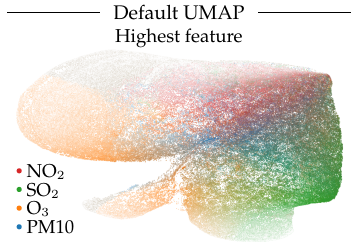}
    \caption{}%
    \label{fig:app:summaries:base}
  \end{subfigure}%
  \begin{subfigure}{0.333\textwidth}
    \centering
    \includegraphics{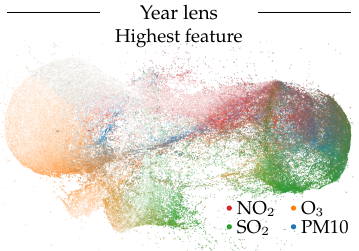}
    \caption{}%
    \label{fig:app:summaries:year}
  \end{subfigure}%
  \begin{subfigure}{0.333\textwidth}
    \centering
    \includegraphics{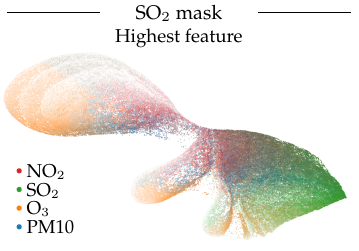}
    \caption{}%
    \label{fig:app:summaries:sulfer}
  \end{subfigure}
  \caption{(Lensed) UMAP embeddings for the Air Quality dataset~\protect\citep{airdata} 
  coloured to summarise the highest feature over the manifold inspired
  by~\citep{silva2015features,thijssen2023explainingprojections}. (\textbf{a})
  Default UMAP (cosine distance 50-nearest-neighbors), (\textbf{b}) a global
  lens over the year dimensions (24 regular segments), and (\textbf{c}) a local
  mask (20 neighbours) over the SO$_2$ dimension. Feature values were normalised
  with a robust z-score enabling direct comparison of their values. The figures
  were created using Datashader's categorical shading that blends hues depending
  on the category values in each pixel~\protect\citep{datashader}.}
  \label{fig:air:summaries}
\end{figure*}

\begin{figure*}[t!]
  \centering
  \begin{subfigure}{0.370\textwidth}
    \centering
    \includegraphics{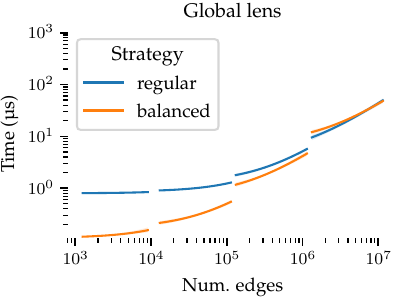}
    \caption{}%
    \label{fig:benchmark:global-lens}
  \end{subfigure}%
  \begin{subfigure}{0.314\textwidth}
    \centering
    \includegraphics{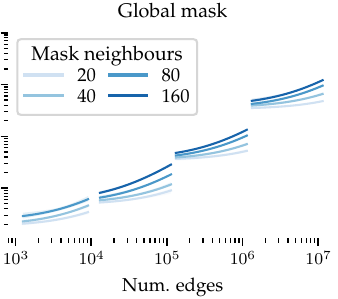}
    \caption{}%
    \label{fig:benchmark:global-mask}
  \end{subfigure}%
  \begin{subfigure}{0.314\textwidth}
    \centering
    \includegraphics{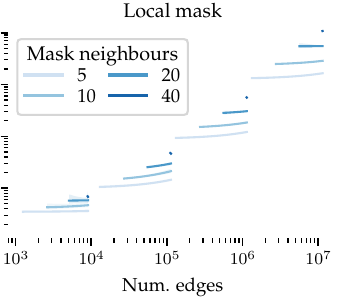}
    \caption{}%
    \label{fig:benchmark:local-mask}
  \end{subfigure}%
  \caption{Benchmark compute times (\si{\micro\second}) excluding the
    embedding step and mask model computation. A linear regression line with its
    95\% interval---relating compute time to the initial UMAP model's edge
    count---is shown for each dataset size (100, 1000, 10.000, 100.000 points),
    lens types, and lens parameter value. (\textbf{a}) The global lens with
    varied discretisation strategy over 3, 6, 12, and 24 segments. (\textbf{b})
    The global mask with 20, 40, 80, and 160 mask neighbours. (\textbf{c}) The
    local mask with 5, 10, 20, and 40 mask neighbours.}%
  \label{fig:benchmark}
\end{figure*}

This use case demonstrates lensed UMAP's ability to deal with a larger dataset.
We adapt an exploratory analysis~\citep{alcaide2015stad} of an air quality
dataset~\citep{airdata} that described several patterns in air quality changes
over time by aggregating the data per week. This pre-processing step effectively
averaged out measurement locations. Here, we show how lensed UMAP explores the
dataset and detects similar patterns while keeping measurement locations
separate.

\subsubsection{Data and Pre-Processing}%
\label{sec:demo:air:data}
The dataset~\citep{airdata} contains daily compound concentration measurements
for several years and locations. Two features with more than 40\% missing values
were removed. Observation with missing values in features with at least 10\%
missing values were also removed. This action reduced the number of data points
from $446.014$ to $181.368$, removing some locations entirely and consecutive
periods from others. The remaining missing values were imputed using their
observation's $5$-nearest neighbours. Finally, a robust z-score was applied to
make the features comparable.

\subsubsection{Exploration Steps}%
\label{sec:demo:air:exploration}
The first exploration step constructed a UMAP model (cosine metric, 50 nearest
neighbours) in \SI{17}{\second} and computed the embedding in \SI{111}{\second}.
The embedding is shown by its edges (Fig.~\ref{fig:air:base-edges}), data
points coloured by year (Fig.~\ref{fig:air:base-year}), and data points
coloured by features (Fig.~\ref{fig:air:base-features}). An equal histogram
normalisation was applied when mapping the features to colours. This technique
preserves value orders but not their magnitudes and avoids outlier values
dominating the colour range~\citep{datashader}. These figures highlight two main
patterns: 1) time appears correlated with the embedding---older observations
appear towards the right side, more recent observations are contained in three
structures on the left and bottom---and 2) three recent structures differ most
in their SO$_2$ values.

A global lens over the observation year was applied to reveal which states exist
each year and how those states progress across years. The lens was configured
with 24 regular segments to retain the edges between equal or consecutive years.
Applying the lens took \SI{29}{\micro\second}, and updating the embedding
required \SI{56}{\second}. The lens effect is visible in
Fig.~\ref{fig:air:year-year}. There appear to be four periods with distinct
structures. Several additional interesting patterns are visible in
Fig.~\ref{fig:air:year-edges} and~\ref{fig:air:year-features}:
\begin{itemize}
  \item PM10 started decreasing from 2003 onward, which may be related to
        vehicle regulations introduced around that time (as cited
        in~\citep{alcaide2015stad}).
  \item Observations before 2008 appear more densely connected, which suggests
        larger differences between non-consecutive years in that period.
  \item The lower and higher NO$_2$ states after 2008 appear connected through
        two arms: one with low and one with high O$_3$ and PM10.
  \item Between 2008 and 2015, the connectivity to previous years occurs through
        observations with relatively high NO$_2$ values. Similar states that
        occur later and states with lower NO$_2$ are located separately.
\end{itemize}

A local mask over the SO$_2$ values (20 neighbours) was applied to inspect that
feature's interaction with the manifold. Applying the mask took
\SI{1.0}{\second}, and the embedding was updated in \SI{49}{\second}. The
resulting embedding is shown in
Fig.~\ref{fig:air:sulfer-edges}-\ref{fig:air:sulfer-features}. These figures
highlight one main pattern: there appear to be multiple slices with different
SO$_2$ values, hinting at some discrete process. Further inspection of the
SO$_2$ values reveals that they are measured in whole \ugmc, and their
distribution is right-tailed. This finding explains explains the slices, as low
SO$_2$ values occur often, and data points with such values are likely to have
20 neighbours with the same value, resulting in few connections to other SO$_2$
values.

A colouring technique that reveals which feature has the highest value along the
manifolds is applied to demonstrate that such visualisation techniques can be
combined with lenses. Our approach approximates the ``value explanation''
from~\cite{silva2015features,thijssen2023explainingprojections} using
Datashader's categorical shading that blends hues for each pixel by the features
means within that pixel~\citep{datashader}. Fig.~\ref{fig:air:summaries} shows
the resulting visualisations that summarise feature behaviour along the
manifold.

\subsubsection{Discussion}%
\label{sec:demo:air:discussion}
The observed lack of connectivity between states after applying a lens can be
caused either by a sufficient change in state or a lack of similar observations
nearby in the lens dimension. Our removal of observations with missing values
contributes to this lack of connectivity because it introduced measurement time
gaps at several locations. The changes in measurement locations over time also
contribute to changes in the observed state. Both factors should be considered
when interpreting the discovered patterns.

While there appear to be spherical structures that reflect seasonal patterns,
they do not correspond to time directly. A location's consecutive measurements
do not move smoothly over the manifold. Instead, day-to-day variations can jump
across the manifold quite wildly. These spherical structures are natural for a
cosine distance metric, as that metric measures the angles between observations
and is sensitive to the relative feature-value compositions.

\subsection{Benchmark}%
\label{sec:demo:benchmark}
Updating the embedding is lensed UMAP's main computational bottleneck, as shown
in Section~\ref{sec:demo:air:exploration}. This step's cost depends only on the
number and weight of the edges being embedded. The local mask's retained number
of edges follows from the specified mask neighbours and the number of data
points. The other two lens types retain more edges as the input model contains
more edges. In our experience, updating an embedding after applying a lens takes
the same order of magnitude time as computing the initial embedding. Removing
too many edges---which would speed up the process---also removes structure,
thereby hindering interpretability. It is possible to accelerate this step with
GPUs, bringing the cost down to seconds for millions of data
points~\citep{nolet2021rapids}.

The global mask has an additional bottleneck: computing the mask model.
Generally, this step tends to be more expensive than computing the original UMAP
model, as more neighbours are needed to balance the mask's strength. As shown in
Section~\ref{sec:demo:air:exploration}, computing a UMAP model on roughly
180.000 data points with 50 neighbours took \SI{12}{\second}. Consequently, this
step will be a noticeable part of the compute time.

Both of these points aside, in this section, we demonstrate the lens types'
computational scaling, excluding both previously mentioned bottlenecks. This
benchmark is not intended to reflect realistic data. Instead, we are interested
in general scaling trends. The benchmark used randomly generated datasets
containing a varying number of points forming 10 clusters around the vertices of
a 10-dimensional hypercube. Then, a UMAP model was computed given a varying
number of neighbours $k$, and the lens types were applied given their parameter
values. All selected parameter values (see Fig.~\ref{fig:benchmark}) were
evaluated on five datasets.

We measured the time required to apply the lenses to the UMAP model, i.e. the
steps described in Section~\ref{sec:algorithm}. For the global mask,
constructing the mask model is excluded from the timing.
Fig.~\ref{fig:benchmark} shows linear regression lines with their 95\% interval
for the compute time (\si{\micro\second}) over the initial UMAP model's edge
count, computed separately for each dataset size, lens type, and indicated lens
parameters.

The global lens' compute time scales linearly with the original model's edge
count, with the balanced strategy being faster at smaller sizes
(Fig.~\ref{fig:benchmark:global-lens}). This difference diminishes as the number
of edges increases. The global mask also scales with the edge count; with a
stronger effect, the more mask neighbours are considered and an additional
effect for the dataset size (Fig.~\ref{fig:benchmark:global-mask}). This pattern
matches the lens type's workload: iterating over the mask and the initial
model's edge union. The local mask appears to scale more with the dataset size
than the initial model's edge count (Fig.~\ref{fig:benchmark:local-mask}). An
increase in edges to process is primarily visible in its interaction with the
number of mask neighbours: the more edges, the stronger the effect of mask
neighbours.

\section{Discussion}%
\label{sec:discussion}
The use cases (Section~\ref{sec:demo}) demonstrated how lens functions enable
analysts to use their domain knowledge in exploring data from multiple
perspectives, leading to different insights. They also demonstrate which lens
type is appropriate for different scenarios:
\begin{itemize}
  \item The global lens is most applicable for separating binary or ordinal
  values, such as the survival state in Section~\ref{sec:demo:nki} and the year
  in Section~\ref{sec:demo:air}. It can also be used for numerical values but is
  limited to one dimension per lens and may split connected components.
  \item The local mask works well on numerical variables, such as CSTA
  expression in Section~\ref{sec:demo:nki} and SO$_2$ values in
  Section~\ref{sec:demo:air}. Its parameters are easy to set, and the mask is
  unlikely to separate connected components.
  \item The global mask is most useful when two related manifolds are available.
  In other cases, the mask manifold's compute cost and indirect nature of the
  parameters limits usability.
\end{itemize} 

The computational costs were reported in the air quality use case
(Section~\ref{sec:demo:air}) and investigated in a synthetic benchmark
(Section~\ref{sec:demo:benchmark}). Generally, updating the embedding
coordinates is the computational bottleneck. In our use cases, this step took
roughly half the time spent computing the initial embedding. Consequently,
applying lenses to larger datasets is not feasible at interactive speeds but
will be quicker than computing the initial model. GPU acceleration can alleviate
this problem~\citep{nolet2021rapids}. Filtering the modelled connectivity---i.e.,
applying the lenses---is much less expensive. The benchmark confirmed this
step's computational cost scales as described in Section~\ref{sec:algorithm}.

\subsection{Validity}%
\label{sec:discussion:validity}
UMAP is based on solid mathematical theory, which gives it
credibility~\citep{mcinnes2018umap}. Lenses break some of the properties UMAP is
designed to maintain by changing what is being modelled from the data's manifold
to the interaction of that manifold and a signal defined on it. While we do not
provide an elaborate theoretical description, we argue that breaking these
properties is justified because lenses uncover the connectivity which a Reeb
graph uses to determine whether two points in a level set are equivalent (i.e.,
in the same connected components). In this interpretation, UMAP provides the
manifold, and the lens types specify how the level sets are defined.

In practice, breaking these properties has consequences in UMAP's sampling-based
embedding process and can reduce the embedding's quality after applying a lens.
For example, the spectral initialisation expects the manifold to contain few
connected connected components and deteriorates when many separate components
are present. In addition, every data point is given at least one attraction
force in every epoch by ensuring it is fully connected to its nearest neighbour.
 
Lenses can break both properties by splitting connected components into smaller
pieces and removing edges between nearest neighbours. The quality of the
resulting embedding is maintained by using the initial model's layout as
initialisation. A reduced attraction force is typically counteracted by
decreasing the repulsion strength. Alternatively, the edge weights can be
normalised after applying a lens---increasing the modelled similarity---or the
lens's strength can be reduced to remove fewer edges. 

We recommend visualising model edges to judge the embedding quality. Long and
overlapping edges indicate too much repulsion and too little attraction. Other
(force-directed) graph layout algorithms can also be used to embed the model
(f.i.,~\cite{Hu2005sfdp,Zhu2020DRGraph,Zhong2023tForce}).

\subsection{Alternatives}%
\label{sec:discussion:alternatives}
Several alternative data exploration techniques have been mentioned in the
present paper. This section compares these approaches to the proposed lens
types, describing their differences and highlighting how the techniques can be
combined. 

The first alternative technique uses colour to summarise feature distributions
across embeddings~\citep{silva2015features, thijssen2023explainingprojections}
(Section~\ref{sec:related_work:enriched_projections}). This approach is not a
direct alternative for lens functions. Instead, it visualises regions where
particular (combinations of) features have stable or extreme values, providing a
figure explaining how regions in the manifold differ. The approach can be
applied with a lens, which we demonstrate in Fig.~\ref{fig:air:summaries}.

The second alternative technique is Mapper, a source of inspiration for our
work. Mapper uses lens functions to visualise a dataset's structure from
configurable perspectives, raising questions to explore and leading to insights.
We attempted to bring this functionality to UMAP in an accessible manner, where
UMAP provides a starting point without needing a lens. The main difference
between both approaches is that Mapper uses clusters while UMAP works with
individual data points. Consequently, it can be easier to estimate how many
points are part of a pattern in a lensed UMAP projection compared to a Mapper
graph. Furthermore, there is no need to inspect and tune clustering behaviours
when using lensed UMAP. Instead, UMAP's $k$-nearest neighbours determine whether
data points are connected.

Finally, we discuss two alternative ways to integrate information with UMAP.
Firstly, some lens dimensions can be added to UMAP's distance metric. Secondly,
lens dimensions can be used to pre-compute a sparse distance matrix, prescribing
which edges UMAP may use. Like lens functions, both approaches increase the
separation between data points that differ in the lens dimensions. Unlike lens
functions, they also decrease the separation between data points with similar
lens values. Computationally, both approaches are more expensive in interactive
exploration workflows as they require recomputing the nearest neighbours for
every lens.

\section{Conclusion}%
\label{sec:conclusion}
The present paper proposed three types of lens functions for UMAP models. Two
use cases demonstrated how the lens types can be use to explore data from
different perspectives, leading to new hypotheses and insights. Lenses are
particularly effective in discovering patterns in a subset of the data, as these
may not be obvious from individual distributions. In addition, lenses can be
based on metadata---i.e., features not included in the distance
metric---providing additional exploration flexibility.

\section*{Acknowledgment}
This work was supported by Hasselt University BOF grant [BOF20OWB33] and KU
Leuven grant STG/23/040.

\section*{Data Availability}
Data is available on-line at \url{https://doi.org/10.5281/zenodo.11193167}.

\bibliography{references.bib}
\end{multicols}
\end{document}